\documentclass{article}

\usepackage[main,final]{neurips_2026}

\usepackage[utf8]{inputenc} 
\usepackage[T1]{fontenc}    
\usepackage{hyperref}       
\usepackage{url}            
\usepackage{booktabs}       
\usepackage{amsfonts}       
\usepackage{nicefrac}       
\usepackage{microtype}      
\usepackage{xcolor}         
\usepackage{graphicx}
\usepackage{amsmath}
\usepackage[table]{xcolor}
\usepackage{wrapfig}
\usepackage{multirow}

\hypersetup{
  colorlinks=true,
  urlcolor=blue 
}

\title{CompactAttention: Accelerating Chunked Prefill with Block-Union KV Selection}

\author{%
  Jiwon Song\quad Dongwon Jo\quad Beomseok Kang\quad Jae-Joon Kim\thanks{Corresponding Author} \\
  Seoul National University \\
  \texttt{\{jiwon.song, dongwonjo, beomseok, kimjaejoon\}@snu.ac.kr} \\
  \url{https://github.com/jiwonsong-dev/CompactAttention}
}

\begin{document}

\maketitle

\begin{abstract}

Chunked prefill has become a widely adopted serving strategy for long-context large language models, but efficient attention computation in this regime remains challenging. Existing sparse attention methods are primarily designed for one-shot prefill and do not translate efficiently to chunked prefill: block-sparse kernels lose efficiency when the query length is limited by the chunk size, while fine-grained pattern search becomes costly when repeated over the accumulated KV cache at every chunk. 
QUOKA, a recent method that directly targets chunked prefill, avoids sparse-kernel overhead but relies on query-subsampled, token-level KV selection, which can miss query-specific KV entries and introduce explicit KV-copy overhead.
To address these limitations, we propose \textbf{CompactAttention}, a chunked-prefill attention mechanism based on \emph{Block-Union KV Selection}.
CompactAttention treats 2D block-sparse masks as KV-selection signals rather than direct sparse-kernel execution plans, and converts them into GQA-aware per-group KV block tables through Q-block union and intra-group union. 
This construction produces block tables that preserve all KV blocks selected by the input masks under paged execution constraints, enabling selected KV blocks to be accessed in place without explicit KV compaction. 
On LLaMA-3.1-8B-Instruct, CompactAttention maintains accuracy close to dense attention on the RULER benchmark while delivering up to 2.79$\times$ attention speedup over dense attention and 1.25$\times$ over FlashPrefill at 128K context length under chunked prefill.



\end{abstract}

\section{Introduction}
\label{sec:intro}

As large language models (LLMs) are increasingly used for long-horizon reasoning, document understanding, code analysis, and agentic workloads, their supported context windows have grown rapidly, reaching hundreds of thousands to even millions of tokens in recent proprietary and open-source models~\cite{gpt5,claudeopus46,gemini3pro,kimik25,deepseekv4}.
Processing such long contexts in a single prefill pass is increasingly impractical.
First, full-sequence attention incurs quadratic compute cost with respect to context length, making one-shot prefill expensive at long contexts.
Second, in online serving systems where prefill and decode requests are batched together, a long prefill pass can stall decode requests, making it difficult to satisfy time-between-token (TBT) service-level objectives (SLOs).

\begin{figure}[t]
    \vspace{-8pt}
    \includegraphics[width=1.0\textwidth]{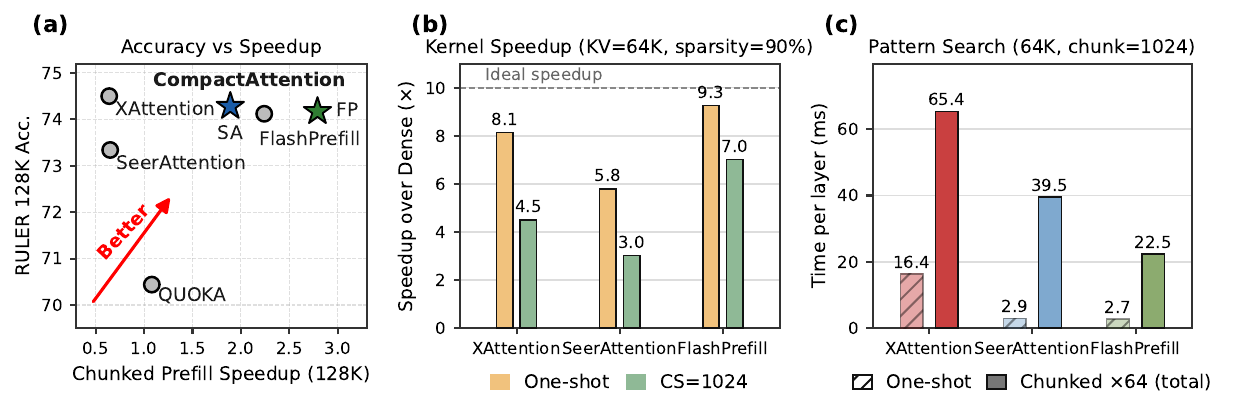}
    \vspace{-20pt}
    \caption{\textbf{(a)} CompactAttention achieves the best accuracy--speedup trade-off. \textbf{(b)} Block-sparse kernels under chunked prefill ($Q \ll KV$) fall far below one-shot and ideal speedups. \textbf{(c)} Pattern search cost accumulates across chunks, with XAttention incurring the highest overhead.}
    \vspace{-8pt}
    \label{fig:scatter_blocksparse}
\end{figure}

Chunked prefill~\cite{sarathi,sarathiserve}, now adopted in major serving frameworks such as vLLM~\cite{vllm} and SGLang~\cite{sglang}, addresses these issues by processing long inputs sequentially in fixed-size chunks, each attending to both its own KVs and the accumulated KV cache from previous chunks.
This makes efficient attention under chunked prefill an increasingly important problem.

The dominant approach to accelerating long-context prefill is block-sparse attention.
Since FlashAttention~\cite{flashattention,flashattention2,flashattention3} operates on blocks of tokens, block-sparse methods~\cite{minference,flexprefill,xattention,seerattention,flashprefill} first estimate which attention blocks are important and then compute only the selected subset of the attention map.
These methods can be effective for one-shot prefill, where the query and key-value lengths are both large enough for sparse execution to amortize irregular memory-access overheads.

However, directly applying block-sparse attention to chunked prefill exposes two limitations.
First, sparse execution becomes inefficient when $Q \ll KV$: block-sparse kernels have too few query blocks to expose sufficient parallelism and amortize irregular access overheads, so the achieved speedup falls far below what the nominal sparsity would suggest.
Second, sparse pattern search must be repeated at every chunk over the accumulated KV cache, making cumulative search overhead a first-order concern and leaving only lightweight pattern-search mechanisms practical.
An alternative is to perform dense attention over a selected subset of KV entries, avoiding block-sparse kernel overhead entirely.

QUOKA~\cite{quoka} is a representative method that directly targets chunked prefill by avoiding sparse kernels and performing dense attention over a reduced set of KV entries selected by a subsampled set of queries.
However, it introduces two limitations.
First, KV entries critical to non-sampled queries can be missed, leading to accuracy degradation on tasks requiring distributed information access.
Second, token-level selection requires explicit KV gathering before attention execution, introducing copy overhead that grows with context length and batch size.

We propose \textbf{CompactAttention}, a chunked-prefill attention mechanism that decouples block-level KV selection from sparse-kernel execution.
The key idea is to separate how KV blocks are selected from how they are executed: CompactAttention reuses lightweight block-sparse pattern search methods for selection, while lowering their 2D masks into GQA-aware KV block tables for zero-copy paged execution.
It converts per-query-block, per-head masks into per-group KV block tables through Q-block union and intra-group union, producing tables that retain all selected KV blocks under paged execution constraints.
These block tables are then passed to a paged attention kernel, which accesses the selected KV blocks in place without explicit KV compaction.
By executing block-level KV selection through a Grouped-Query Attention~\cite{gqa} (GQA)-aware dense paged-attention backend, CompactAttention avoids both the kernel inefficiency of block-sparse attention and the copy overhead of token-level KV selection.

We evaluate CompactAttention on long-context LLMs under chunked prefill.
As summarized in Figure~\ref{fig:scatter_blocksparse}(a), CompactAttention achieves the best accuracy--speedup trade-off among all baselines on the RULER benchmark, maintaining accuracy close to dense attention while delivering up to 2.79$\times$ speedup at 128K context length on H200.
These results show that block-union KV table construction and zero-copy paged execution directly address the execution bottlenecks of existing chunked-prefill attention methods.

\section{Motivation}
\label{sec:motivation}

\begin{figure}[t]
    \includegraphics[width=1.0\textwidth]{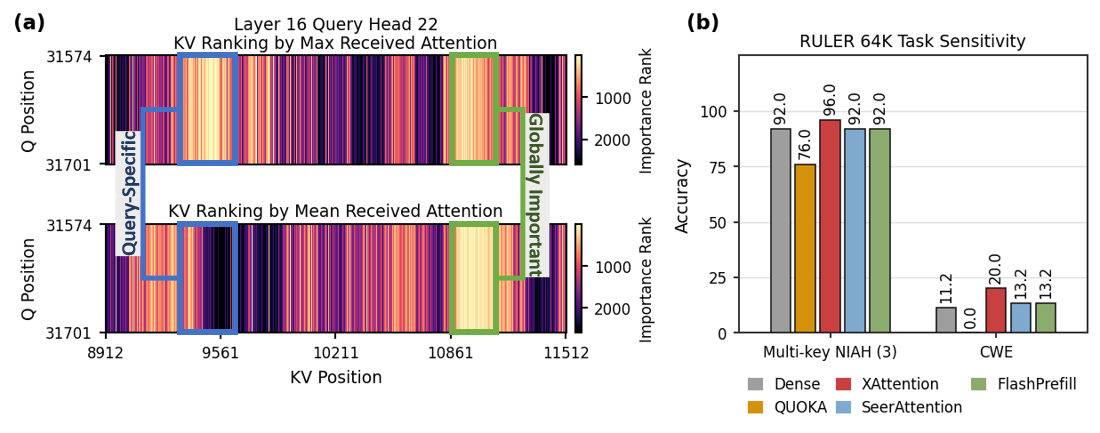}
    \vspace{-20pt}
    \caption{\textbf{(a)} KV-position rankings obtained by aggregating the attention each KV position receives from query positions in the shown window.
    Mean received attention ranks KV positions by their average received attention across all queries in the shown window, emphasizing globally important KV positions. Max received attention ranks KV positions by the largest attention received from any query within the window, exposing query-specific KV positions that may be important to only a few queries.
    Brighter colors indicate higher importance; highlighted regions show query-specific and globally important KV positions.
    \textbf{(b)} QUOKA degrades on tasks requiring distributed information access, while block-sparse methods remain close to dense attention.}
    \vspace{-8pt}
    \label{fig:motivation}
\end{figure}

\subsection{Block Sparse Attention under Chunked Prefill}
\label{subsec:motivation_block_sparse}

Block-sparse attention has been the dominant paradigm for accelerating long-context prefill.
These methods identify input-dependent sparse patterns for each attention head and compute only a selected subset of attention tiles, skipping tiles that are predicted to be unimportant.
By exploiting sparsity while preserving most of the relevant attention computation, they can achieve substantial speedups with accuracy close to dense attention in one-shot prefill. 
However, applying these methods directly to chunked prefill exposes two key limitations.

\paragraph{Kernel Inefficiency.}
In chunked prefill, the query length at each iteration is limited to the chunk size, typically a few hundred to a thousand tokens in multi-request serving batches, while the key-value length grows as chunks accumulate. 
This $Q \ll KV$ regime differs substantially from one-shot prefill, where $Q=KV$. 
As shown in Figure~\ref{fig:scatter_blocksparse}(b), at the same KV length of 64K and 90\% sparsity, block-sparse kernels achieve speedup much closer to the ideal (10$\times$) value under one-shot prefill ($Q=64K$) than under chunked prefill ($Q=1024$).
This gap arises because block-sparse kernels rely on sufficiently large attention tiles to amortize the fixed overhead of sparse mask interpretation and irregular memory access. When the query sequence is short, the number of active query blocks is small, and this overhead dominates over the savings from skipping attention tiles.

\paragraph{Pattern Search Overhead.}
Another challenge is the cost of finding input-dependent sparse patterns.
Reducing this cost has been a central focus of block-sparse attention research.
For example, XAttention~\cite{xattention} substantially reduces scoring overhead compared with earlier fine-grained methods such as MInference~\cite{minference} and FlexPrefill~\cite{flexprefill}.
However, as shown in Figure~\ref{fig:scatter_blocksparse}(c), chunked prefill amplifies the cost of any online pattern search because scoring must be repeated at every chunk over the accumulated KV cache.
This makes chunked prefill sensitive to the choice of pattern search method. 
Among existing block-sparse methods, lightweight selectors such as SeerAttention~\cite{seerattention} and FlashPrefill~\cite{flashprefill} are therefore the most practical choices for this regime, although their cumulative search overhead remains higher under chunked prefill than under one-shot prefill.
This constraint motivates a selector-agnostic execution design that can use practical lightweight methods today while remaining compatible with faster search mechanisms in the future.


\subsection{Limitations of Query-Subsampled Direct KV Selection}
\label{subsec:motivation_query_subsampled}


Rather than using a sparse attention kernel, QUOKA~\cite{quoka} subsamples a subset of query tokens from the current chunk to score the importance of cached KV entries, and performs dense attention over the selected KV tokens.
However, query-subsampled selection has an inherent coverage limitation.
As illustrated in Figure~\ref{fig:motivation}(a), we rank each KV position by aggregating the attention it receives from query positions in the shown window.
Mean-attention ranking highlights globally important KV positions that receive attention broadly across queries, while max-attention ranking reveals query-specific positions that receive strong attention from only a small subset of queries.
Because only sampled queries participate in QUOKA's KV scoring, such query-specific KV entries may be missed when their corresponding queries are not selected as evaluators.
As shown in Figure~\ref{fig:motivation}(b), this coverage limitation appears on RULER tasks that require distributed information access.
QUOKA degrades noticeably on Multi-key NIAH-3 and CWE, while block-sparse methods remain close to dense attention by evaluating all query blocks.
Furthermore, unlike block-sparse methods that operate on contiguous blocks of tokens, QUOKA selects KV entries at token granularity.
The selected KVs must therefore be gathered into a reduced KV tensor before attention execution, introducing explicit copy overhead that grows with context length and batch size.

These observations motivate a different design for chunked prefill. 
An effective mechanism should cover all query blocks, avoid sparse-kernel inefficiency in the short-query regime, and select KVs at block granularity for direct access without explicit compaction.
\textbf{CompactAttention} is designed around these requirements by decoupling block-level KV selection from attention execution.
This design is a trade-off rather than a free improvement: by keeping every KV block selected by any query block or head within an execution group, CompactAttention trades fine-grained sparsity for coverage of query-specific KV and for regular, copy-free execution.

\section{CompactAttention}
\label{sec:method}

\subsection{Overview}

\begin{figure}[t]
    \includegraphics[width=1.0\textwidth]{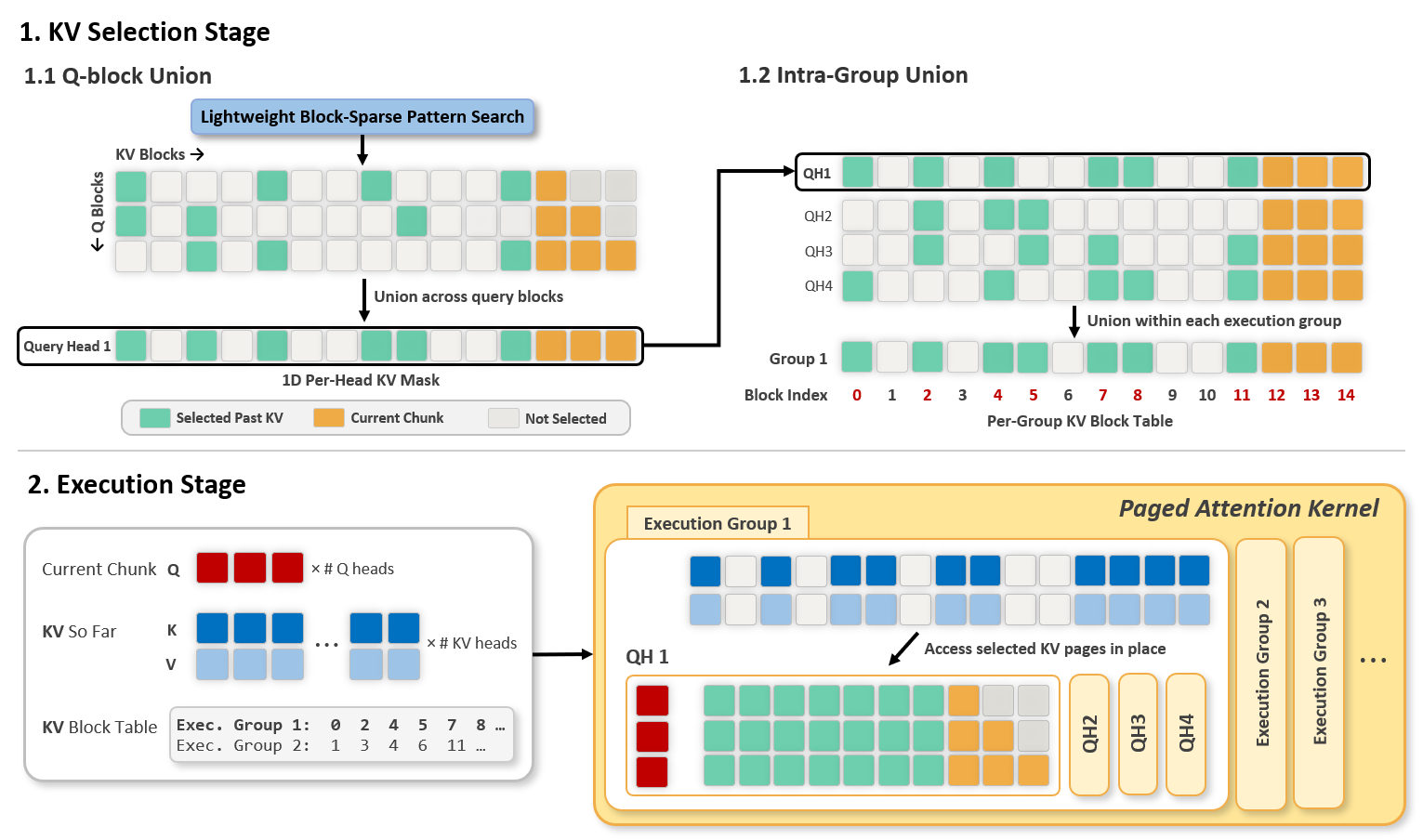}
    \vspace{-12pt}
    \caption{\textbf{Overview of CompactAttention.} The KV selection stage converts a 2D per-head block mask into per-group KV block tables through Q-block union and intra-group union. The execution stage passes these block tables to a paged attention kernel, which accesses selected KV pages in place without explicit KV compaction.
    }
    \vspace{-8pt}
    \label{fig:overview}
\end{figure}

CompactAttention is a chunked prefill-aware attention mechanism that decouples sparse KV selection from execution, as illustrated in Figure~\ref{fig:overview}. 
It can be combined with any lightweight block-sparse pattern search method that provides block-level importance estimates with low per-chunk overhead.
Given block-level importance scores, CompactAttention proceeds in two stages: \textit{selection} and \textit{execution}.

In the \textit{selection} stage, it converts per-head sparse masks into compact KV block tables through two union operations.
First, it applies Q-block union across query blocks within the current chunk, which is necessary because dense paged attention consumes a single KV block list for the query blocks executed together rather than separate decisions for each $(Q\text{-block}, KV\text{-block})$ tile.
Second, it applies intra-group union across query heads that are executed together, producing one KV block table per group.
In the \textit{execution} stage, each per-group block table is passed to a paged attention kernel, which accesses the selected KV blocks in place without copying them into a separate buffer.
This zero-copy execution avoids the copy overhead of token-level KV selection methods while leveraging optimized dense paged-attention kernels.
The current chunk is always kept fully open to preserve causal attention semantics.
Details of the \textit{selection} process and the \textit{execution} process are described in Section~\ref{subsec:selection} and Section~\ref{subsec:paged_attention}, respectively.

\subsection{KV Selection: Block-Union KV Table Construction}
\label{subsec:selection}

CompactAttention's selection stage accepts any block-sparse pattern search method that produces a 2D per-head block mask with sufficiently low per-chunk overhead.
In this work, we instantiate CompactAttention with two lightweight pattern search methods: SeerAttention (SA)~\cite{seerattention}, a learned attention-pattern predictor, and FlashPrefill (FP)~\cite{flashprefill}, a training-free method based on max-threshold dynamic thresholding.

Let \(M_{b,h,i,j} \in \{0,1\}\) denote the 2D block-sparse mask produced by the pattern search method for batch \(b\), query head \(h\), query block \(i\), and KV block \(j\). 
Existing block-sparse methods use this mask directly as a sparse-kernel execution plan. CompactAttention instead converts it into a KV block table that can be consumed by dense paged attention.

CompactAttention first applies \emph{Q-block union} across query blocks:
\[
\bar{M}_{b,h,j} = \bigvee_i M_{b,h,i,j}.
\]
This produces a single 1D KV block mask per query head. The union is required because dense paged attention consumes one KV block list for the query blocks executed together. 
CompactAttention then applies \emph{intra-group union} across query heads that share a KV block table:
\[
G_{b,g,j} = \bigvee_{h \in \mathcal{H}(g)} \bar{M}_{b,h,j},
\]
where \(\mathcal{H}(g)\) denotes the set of query heads in an execution group \(g\), which is a KV group by default.
The resulting per-group page table is
\[
T_{b,g} = \{j \mid G_{b,g,j}=1\}
=
\{j \mid \exists h \in \mathcal{H}(g), \exists i,\; M_{b,h,i,j}=1\}.
\]

This block-union construction is coverage-preserving with respect to the input block-sparse mask: a KV block selected by any query block under any query head in the group remains selected in the resulting page table.
Moreover, under the constraint that all query blocks and all query heads within an execution group share a single KV block table, \(T_{b,g}\) is the minimal table that preserves this coverage.
Any KV block outside \(T_{b,g}\) is not selected by any query block or query head in the input mask, and can therefore be excluded without violating coverage preservation.
Thus, the two union operations are not merely post-processing.
They lower per-query-block, per-head sparse masks into GQA-aware paged KV tables, an executable representation for dense paged attention.
This lowering preserves all KV blocks selected by the original 2D mask while enabling group-wise zero-copy execution.

The two union operations reduce sparsity compared to the original 2D block-sparse mask, because a KV block is retained if it is selected by any query block or query head in the execution group. 
However, we observe that this sparsity reduction can be compensated by using a more aggressive pattern search for the initial 2D mask while still preserving accuracy after union.
As shown in Section~\ref{subsec:speedup}, CompactAttention still achieves higher attention speedup than the corresponding block-sparse baselines, indicating that the execution advantage of dense paged attention outweighs the sparsity loss in practice.

For models with large GQA groups, applying intra-group union across the full group can cause excessive sparsity loss.
We therefore split each KV group into smaller execution groups and apply intra-group union independently within each group.
In our implementation, we use a subgroup size of four query heads as a conservative default; further details are provided in Appendix~\ref{subsec:subkvgroup}. 

\subsection{Execution: Zero-Copy Paged Attention}
\label{subsec:paged_attention}

CompactAttention executes the selected KV blocks using a paged dense-attention backend while avoiding explicit K/V compaction. 
The key requirement is to expose each selected KV block as a page that the backend can access directly, even when different groups use different block tables.
Thus, the block-union table produced in Section~\ref{subsec:selection} must be represented as metadata over the original KV cache rather than as a newly materialized compact KV tensor.

\begin{wrapfigure}{r}{0.45\columnwidth}
  \centering
  \vspace{-12pt}
  \includegraphics[width=0.45\columnwidth]{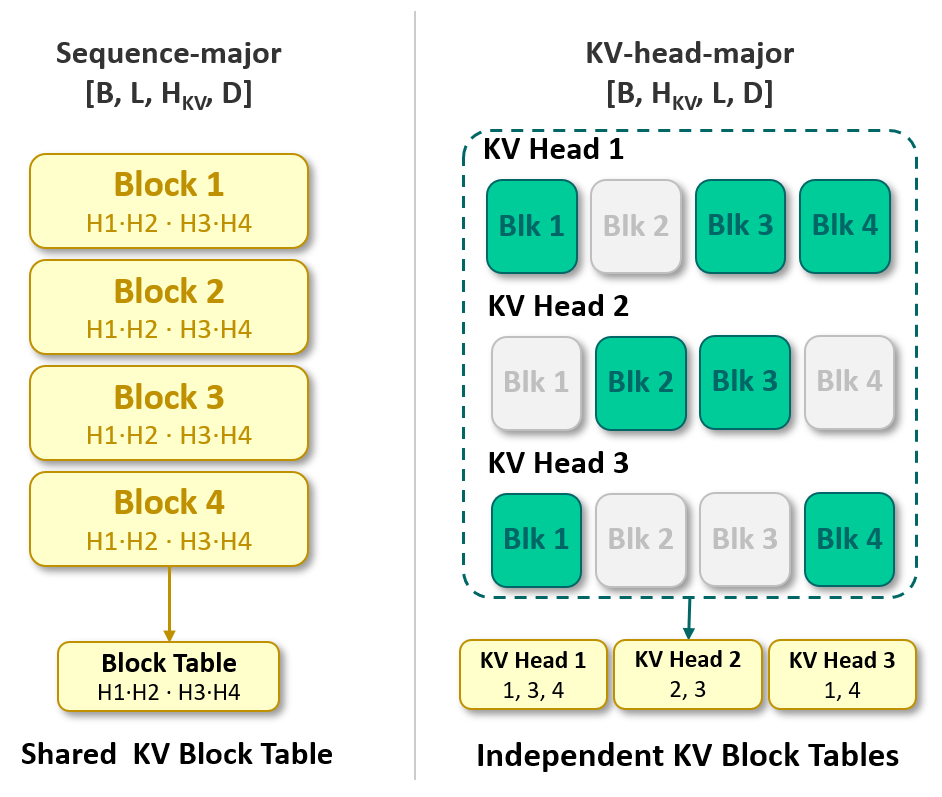}
  \caption{\textbf{KV cache layout comparison.}
  Sequence-major layout forces KV heads to share one block table, preventing independent block selection.
  KV-head-major layout exposes each KV-head block as a page, enabling independent KV block tables without copying K/V payloads.}
  \vspace{-4pt}
  \label{fig:cache_layout}
\end{wrapfigure}

As illustrated in Figure~\ref{fig:cache_layout}, a sequence-major KV cache layout $[B, L, H_{kv}, D]$ forces all KV heads to share the same block table. This is insufficient for CompactAttention because its block tables are group-dependent.
CompactAttention therefore stores the accumulated KV cache in a KV-head-major layout, $[B, H_{kv}, L, D]$, where each $(\text{batch}, \text{KV head}, \text{block})$ triple corresponds to a contiguous $[\text{block size}, D]$ memory region.

This layout is not merely an implementation detail: it turns selected KV blocks into metadata-addressable pages.
CompactAttention constructs a ragged page list independently for each $(\text{batch}, \text{group})$ row, passing only metadata---\texttt{kv\_indptr} and \texttt{kv\_indices}---to the paged attention kernel while reusing the original K/V payloads in place.
Further implementation details are provided in Appendix~\ref{subsec:zerocopy}.

This zero-copy design avoids explicit compaction into a newly allocated dense buffer, whose memory bandwidth overhead grows with context length, batch size, and the number of selected KV blocks. 
Since CompactAttention uses a standard paged dense-attention backend, improvements to dense attention kernels can be adopted without changing the selection stage.

\section{Experiments}
\label{sec:experiments}

\subsection{Experimental Setup}

\paragraph{Models.} 
We evaluate on two open-source models. LLaMA-3.1-8B-Instruct~\cite{llama3} is a dense LLM with a 128K-token context window. 
Qwen3-30B-A3B-Instruct-2507~\cite{qwen3} is a Mixture-of-Experts LLM with a 256K-token context window. 
Both models use Grouped-Query Attention (GQA).
For accuracy evaluation, we use two long-context benchmarks: RULER~\cite{ruler} and LongBench V2~\cite{longbenchv2}.

\paragraph{Compared Methods.} 
We compare CompactAttention against several baselines. 
For dense attention, we use  FlashInfer 0.6.9~\cite{flashinfer} with FlashAttention-2~\cite{flashattention2} and FlashAttention-3~\cite{flashattention3} backends depending on the device. 
For block-sparse attention, we include SeerAttention~\cite{seerattention} with block size 64, XAttention~\cite{xattention} with block size 128, and FlashPrefill~\cite{flashprefill} with block size 128. 
QUOKA~\cite{quoka} is the most directly comparable baseline for chunked-prefill KV selection; it selects KV entries via query subsampling and executes attention with a dense kernel.

CompactAttention uses the FlashInfer infrastructure as the paged attention execution backend, but supplies per-group KV block tables as page metadata to attend only to the selected KV blocks.
CompactAttention-SA uses the pre-trained SeerAttention gate released for LLaMA-3.1-8B-Instruct without modification.
CompactAttention-FP applies FlashPrefill's training-free thresholding and requires no model-specific adaptation, enabling evaluation on both models. 
In both cases, CompactAttention adopts the block size of the corresponding block-sparse attention method.
As no pre-trained SeerAttention gate is available for Qwen3-30B-A3B-Instruct-2507, SeerAttention and CompactAttention-SA are evaluated only on LLaMA-3.1-8B-Instruct.

For QUOKA, we use the fixed 25\% KV budget from the original paper.
For all other sparse methods, we set sparsity hyperparameters independently for each method and model to select accuracy-preserving operating points.
For XAttention, we construct a head-wise threshold table for each evaluated model using the official implementation.
For LLaMA-3.1-8B-Instruct, we use threshold \(=3\mathrm{e}{-4}\) for SeerAttention, threshold \(=5\mathrm{e}{-4}\) for CompactAttention-SA, \(\alpha=0.01\) for FlashPrefill, and \(\alpha=0.06\) for CompactAttention-FP.
For Qwen3-30B-A3B-Instruct-2507, we use \(\alpha=0.02\) for FlashPrefill and \(\alpha=0.12\) for CompactAttention-FP.

\paragraph{Environment.} 
We measure attention latency on two NVIDIA GPUs. 
The RTX PRO 6000 features 96~GB of GDDR7 memory and is based on the Blackwell microarchitecture (SM120), supporting FlashAttention-2. 
The H200 SXM provides 141~GB of HBM3e memory and is based on the Hopper microarchitecture (SM90), enabling FlashAttention-3 with Hopper-specific optimizations.

\subsection{Speedup}
\label{subsec:speedup}

\begin{figure}[t]
    \includegraphics[width=1.0\textwidth]{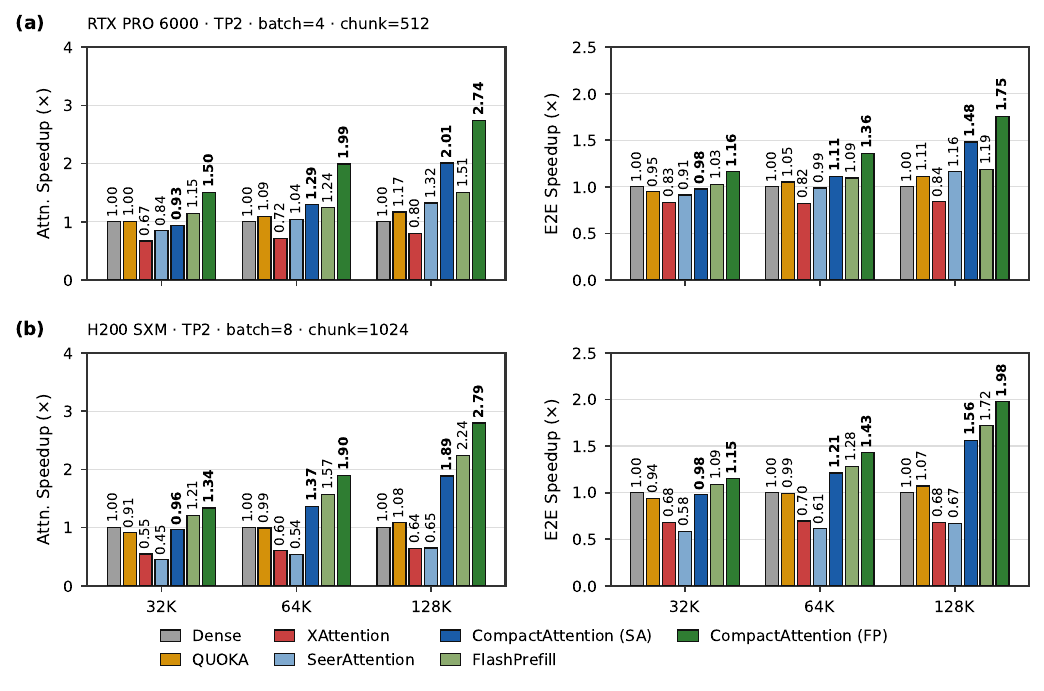}
    \vspace{-22pt}
    \caption{\textbf{Attention and end-to-end speedup under chunked prefill.}
    We report speedup over dense attention on LLaMA-3.1-8B-Instruct across context lengths.
    \textbf{(a)} RTX PRO 6000 GPUs with TP=2, batch size 4, and chunk size 512.
    \textbf{(b)} H200 SXM GPUs with TP=2, batch size 8, and chunk size 1024.
    CompactAttention achieves the largest gains at long context lengths, and the attention-level improvements translate into end-to-end latency reductions.
    }
    \vspace{-12pt}
    \label{fig:speedup}
\end{figure}

Figure~\ref{fig:speedup} reports attention-level and end-to-end speedup over the dense-attention baseline on LLaMA-3.1-8B-Instruct under chunked prefill, where end-to-end latency measures total wall-clock time for chunked prefill.
We evaluate RTX PRO 6000 (TP=2, batch size 4, chunk size 512) and H200 SXM (TP=2, batch size 8, chunk size 1024).
Raw LLaMA latency values and additional Qwen3-30B-A3B-Instruct-2507 speedup results are provided in Appendix~\ref{appendix:additional_latency}. 

QUOKA achieves only limited speedup at long context lengths, as the token-level gather-and-pack overhead offsets the gain from attending to fewer tokens.
XAttention and SeerAttention are often slower than dense attention, reflecting repeated pattern search overhead and inefficient block-sparse execution in the \(Q \ll KV\) regime.
FlashPrefill is the strongest block-sparse baseline, benefiting from lightweight pattern search and optimized block-sparse execution.

CompactAttention-SA and CompactAttention-FP show increasing speedup as context length grows.
On H200 at 128K, CompactAttention-FP reaches 2.79$\times$ attention speedup and 1.98$\times$ end-to-end speedup over the dense-attention baseline.
Against FlashPrefill, the strongest baseline, the gain is more modest: 1.25$\times$ in attention latency and 1.15$\times$ end-to-end at 128K. 
Both CompactAttention variants improve over their corresponding block-sparse baselines at long context lengths, showing that dense paged execution of the unioned KV set outweighs the sparsity reduction from block union.


\subsection{Accuracy}
\label{subsec:accuracy}

\begin{table*}[t]
\centering
\caption{\textbf{RULER accuracy across context lengths.} All methods are evaluated with chunk size 1024.}
\label{tab:ruler}
\scalebox{0.92}{
\renewcommand{\arraystretch}{0.95}
\setlength{\tabcolsep}{7.5pt}
\begin{tabular}{lcccc|cccc}
\toprule
& \multicolumn{4}{c|}{LLaMA-3.1-8B-Instruct} & \multicolumn{4}{c}{Qwen3-30B-A3B-Instruct-2507} \\
\cmidrule(lr){2-5} \cmidrule(lr){6-9}
Method & 32K & 64K & 128K & Avg & 32K & 64K & 128K & Avg \\
\midrule
\rowcolor{gray!8}
Dense                 & 88.23 & 83.52 & 76.59 & 82.78 & 93.01 & 90.56 & 87.81 & 90.46 \\
QUOKA                 & 83.52 & 79.99 & 70.44 & 77.98 & 81.96    & 79.57    & 79.03    & 80.19    \\
XAttention            & 89.16 & 83.27 & 74.50 & 82.31 & 92.29    & 88.92    & 88.13    & 89.78    \\
SeerAttention         & 90.23 & 83.82 & 73.34 & 82.46 & \multicolumn{4}{c}{--} \\
\rowcolor{blue!8}
\textbf{CompactAttention}-SA & 88.92 & 84.12 & 74.28 & 82.44 & \multicolumn{4}{c}{--} \\
FlashPrefill          & 88.99 & 83.94 & 74.12    & 82.35    & 91.55    & 87.73    & 86.77    & 88.68    \\
\rowcolor{blue!8}
\textbf{CompactAttention}-FP & 88.77 & 83.96 & 74.17 & 82.30 & 92.41    & 88.43    & 87.26    & 89.37    \\
\bottomrule
\end{tabular}
}
\end{table*}

\paragraph{RULER.} 
Table~\ref{tab:ruler} reports RULER accuracy across context lengths for all methods.
QUOKA consistently underperforms dense attention across both models and context lengths.
This is consistent with the coverage limitation discussed in Section~\ref{subsec:motivation_query_subsampled}: query-subsampled KV scoring can miss KV entries that are important to unsampled query positions, leading to degradation on tasks requiring distributed information access.
Block-sparse attention methods---XAttention, SeerAttention, and FlashPrefill---remain close to dense attention, suggesting that block-level selectors that evaluate all query blocks better preserve the relevant attention computation.

CompactAttention-SA and CompactAttention-FP exhibit similar accuracy to their corresponding block-sparse baselines because they reuse the same block-level pattern search and preserve the selected blocks through Q-block union and intra-group union.
Thus, CompactAttention largely preserves the selection quality of block-sparse methods while avoiding their sparse-kernel inefficiency under chunked prefill.

\begin{figure}[t]
    \vspace{-8pt}
    \includegraphics[width=1.0\textwidth]{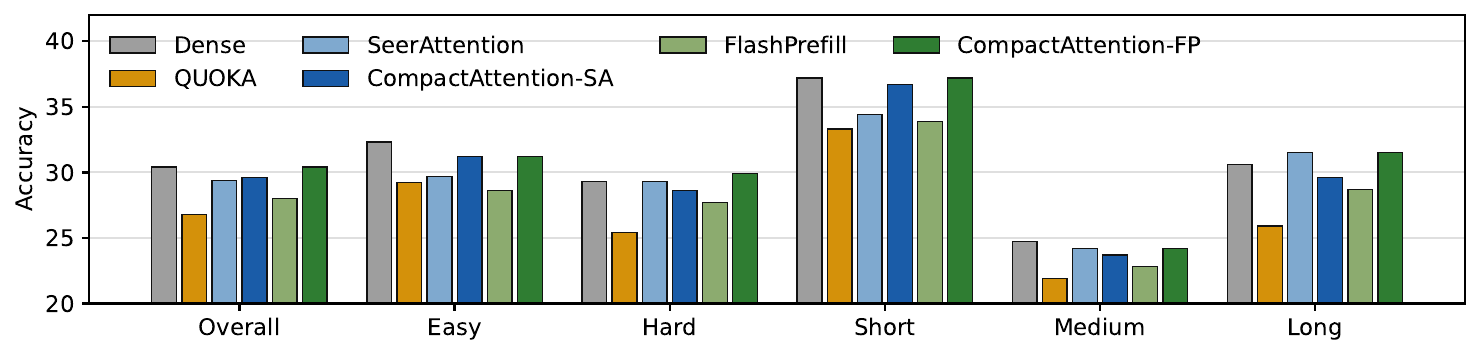}
    \vspace{-16pt}
    \caption{\textbf{LongBench V2 accuracy on LLaMA-3.1-8B-Instruct with chunk size 1024.}
    CompactAttention variants remain close to dense attention across difficulty levels and context-length groups, while QUOKA degrades more noticeably, especially on Hard samples.}
    \vspace{-6pt}
    \label{fig:longbench}
\end{figure}

\paragraph{LongBench V2.}
Figure~\ref{fig:longbench} reports LongBench V2 accuracy on LLaMA-3.1-8B-Instruct with chunk size 1024 across difficulty levels and context-length groups.
The favorable accuracy trend of CompactAttention also extends to LongBench V2, which requires deeper understanding and reasoning over long contexts.
Both CompactAttention variants remain close to dense attention, while QUOKA degrades noticeably, particularly on Hard samples.
Although SeerAttention achieves comparable accuracy, Section~\ref{subsec:speedup} shows that CompactAttention provides a more favorable accuracy–efficiency trade-off under chunked prefill.

\subsection{Ablation Studies}
\label{subsec:ablation}

We focus ablations on CompactAttention-FP (CA-FP), the training-free instantiation applicable to both evaluated models.

\begin{figure}[t]
    \includegraphics[width=1.0\textwidth]{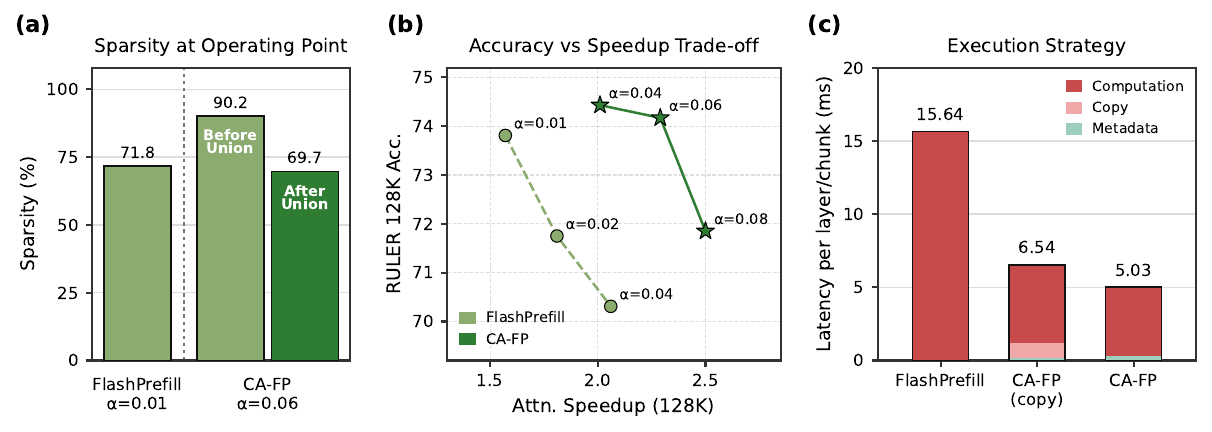}
    \vspace{-16pt}
    \caption{
    \textbf{(a)} Sparsity at the selected operating point.
    \textbf{(b)} Accuracy--speedup trade-off under \(\alpha\) sweep on RULER 128K (RTX PRO 6000, TP=2, batch size 4, chunk size 1024).
    \textbf{(c)} Execution-only ablation at matched sparsity using the same unioned block mask (RTX PRO 6000, 128K, batch size 4, chunk size 512).
    \vspace{-12pt}
}
    \label{fig:ablation}
\end{figure}

\paragraph{Sparsity Analysis.}
Figure~\ref{fig:ablation}(a) compares the sparsity of FlashPrefill and CompactAttention-FP at the selected operating point on RULER  128K.
FlashPrefill with \(\alpha=0.01\) achieves 71.8\% sparsity.
CompactAttention-FP (CA-FP) uses a more aggressive initial FlashPrefill-style mask with \(\alpha=0.06\), whose sparsity decreases from 90.2\% to 69.7\% after Q-block union and intra-group union.
Thus, CompactAttention-FP reaches comparable executed sparsity to FlashPrefill while preserving the selected blocks required by the block-union table.
Post-union keep ratios vary across layers and execution groups, and a few become nearly dense, but these account for a small share of the executed attention at long context; Appendix~\ref{app:sparsity} reports the per-stage, distributional, and cross-task analysis.

\paragraph{Pattern-Search Aggressiveness.}
Figure~\ref{fig:ablation}(b) evaluates the accuracy--speedup trade-off under different \(\alpha\) values on RULER 128K.
We sweep method-specific \(\alpha\) values for FlashPrefill and CompactAttention-FP because block union changes the executed sparsity after the initial FlashPrefill-style mask is generated.
CompactAttention-FP remains favorable in the high-accuracy operating region, achieving higher attention speedup than FlashPrefill at comparable or higher accuracy.

\paragraph{Execution Strategy.}
Figure~\ref{fig:ablation}(c) isolates the effect of execution strategy using the same unioned block mask at matched sparsity. The block-sparse variant executes this mask directly with a block-sparse kernel. 
CompactAttention-FP (Copy) gathers the selected KV blocks into a contiguous buffer before invoking dense attention, while CompactAttention-FP represents the same selected blocks as paged-attention metadata and accesses the original KV cache in place.

We separately report metadata overhead for each execution path. 
CompactAttention-FP incurs metadata overhead from constructing per-group block tables and planning paged attention execution,
whereas CompactAttention-FP (Copy) incurs explicit KV-copy overhead from materializing selected blocks into a compact buffer. 
Since all variants use the same selected KV blocks, this ablation isolates the effect of execution.
CompactAttention-FP achieves the lowest latency despite its metadata cost, showing that in-place paged execution is more efficient than either sparse-kernel execution or explicit KV compaction.

\section{Limitations}
\label{sec:limitations}

CompactAttention inherits the quality of the underlying block-sparse pattern search method: blocks missed by the input mask cannot be recovered by block-union KV selection. 
It also trades sparsity for coverage, since Q-block union and intra-group union retain any KV block selected by any query block or query head within an execution group. 
As a result, the executed sparsity is not set by an explicit budget but is controlled indirectly through the selector hyperparameters and execution-group partitioning, and individual layers or execution groups can become nearly dense.
While dense paged execution outweighs this sparsity loss in our evaluated settings, the trade-off may vary with model architecture, context length, chunk size, sparsity hyperparameters, and execution-group partitioning. 
CompactAttention is most effective when the accumulated KV cache is large enough to amortize pattern search and metadata construction overheads. 
Consistent with this behavior, our end-to-end latency gains become more pronounced as context length increases.
Conversely, at short contexts or small batch sizes these overheads may not be amortized, and CompactAttention can be slower than dense attention or FlashPrefill.
In addition, our implementation builds on HuggingFace Transformers and FlashInfer; integrating CompactAttention into serving frameworks such as vLLM or SGLang requires exposing per-execution-group block tables and a KV-head-major cache layout.

\section{Conclusion}
\label{sec:conclusion}

We presented CompactAttention, a chunked-prefill attention mechanism based on Block-Union KV Selection. 
CompactAttention treats 2D block-sparse masks as KV-selection signals rather than direct sparse-kernel execution plans, converts them into per-group KV block tables through Q-block union and intra-group union, and executes the selected KV blocks in place with zero-copy paged attention.
Across RULER and LongBench V2, CompactAttention maintains accuracy close to dense attention while improving attention latency under chunked prefill, reaching up to 2.79$\times$ speedup over the dense-attention baseline and 1.25$\times$ over FlashPrefill  at 128K context length on H200.
These results demonstrate that the main bottleneck of sparse attention under chunked prefill is not only which KV blocks to select, but also how the selected blocks are executed.
By making sparse selection compatible with efficient dense paged attention kernels, CompactAttention offers a practical path toward faster long-context LLM serving.








\def\bibfont{\small}

\bibliographystyle{unsrt} 
\medskip

\bibliography{references}

\newpage

\appendix

\section{Related Work}
\label{sec:related_work}

\subsection{Chunked Prefill}
\label{subsec:related_chunked_prefill}

Chunked prefill was first proposed by Sarathi~\cite{sarathi}, which splits prefill requests into equal-sized chunks and interleaves them with decode iterations to improve GPU utilization and decode throughput.
Sarathi-Serve~\cite{sarathiserve} built on this idea to directly address the throughput-latency tradeoff, introducing stall-free scheduling that allows new requests to join a running batch without pausing ongoing decodes, simultaneously improving TBT latency and throughput.
Chunked prefill has since been adopted as the default scheduling strategy in major serving frameworks including vLLM~\cite{vllm} and SGLang~\cite{sglang}.

\subsection{Sparse Attention for Long-Context Prefill}
\label{subsec:related_sparse_attention}

A line of work accelerates prefill by skipping unimportant attention blocks. 
MInference~\cite{minference} classifies each attention head's dominant pattern offline and applies the corresponding sparse computation. 
FlexPrefill~\cite{flexprefill} dynamically selects sparse patterns and block indices per head and input via online scoring. XAttention~\cite{xattention} proposes a lightweight scoring mechanism based on anti-diagonal attention values to reduce the overhead of block selection.
FlashPrefill~\cite{flashprefill} further reduces pattern search overhead and achieves high sparsity through fused dynamic thresholding. 
Taking a learning-based approach, SeerAttention~\cite{seerattention} trains a lightweight AttnGate via self-distillation to predict block-level attention pattern while keeping original model weights frozen, achieving high sparsity with low pattern search overhead.
Going further, MoBA~\cite{moba} and NSA~\cite{deepseeknsa} integrate sparsity directly into the model through pretraining or continued training.

These methods succeed in achieving substantial speedup on one-shot prefill without substantially compromising accuracy. 
However, none of them address the kernel inefficiency or per-chunk scoring overhead that arise specifically under chunked prefill.

\subsection{Query-Dependent KV Selection}
\label{subsec:related_query_dependent}

A related line of work reduces the KV cache retained after prefill by selecting only the entries deemed important for decoding. SnapKV~\cite{snapkv} identifies important KV entries using the most recent query tokens as evaluators, discarding the rest before decoding begins. 
Quest~\cite{quest} dynamically selects relevant KV pages at the page level in a query-aware manner at every decoding step.
QUOKA~\cite{quoka} adapts this idea to chunked prefill, sampling a representative subset of query tokens from each chunk as evaluators to score KV importance and performing dense attention over the selected entries.
However, as discussed in Section~\ref{sec:motivation}, query subsampling introduces a structural coverage limitation and token-level selection incurs explicit copy overhead, motivating the block-level selection and zero-copy execution design of CompactAttention.

\newpage

\section{Implementation Details}
\label{sec:implementation}

\subsection{Sub-KV-Group Union}
\label{subsec:subkvgroup}

For models with GQA ratio greater than 4:1 (e.g., Qwen3-30B-A3B-Instruct-2507 with 8:1), applying intra-group union across the full KV group causes excessive sparsity loss.
CompactAttention therefore partitions each KV group into subgroups of four query heads and treats each subgroup as an execution group. 
We refer to this variant as sub-KV-group union. 
We use a subgroup size of 4 as a conservative default rather than a tuned optimum. Smaller subgroups preserve more sparsity but increase the number of block tables to construct, so the best size depends on the context length and chunk size.

As shown in Table~\ref{tab:qwen_subgroup}, sub-KV-group union substantially lowers the keep ratio compared with full KV-group union while maintaining comparable accuracy and the same zero-copy execution interface.


\begin{table}[h]
\centering
\caption{
Subgroup-size trade-off on Qwen3-30B-A3B-Instruct-2507 at 128K context length. Latency is measured on H200 SXM GPUs with TP=2 and batch size 4. Subgroup size 8 corresponds to full KV-group union.}
\label{tab:qwen_subgroup}
\scalebox{0.95}{
\begin{tabular}{cccccc}
\toprule
Chunk size & $\alpha$ & Subgroup size & Keep ratio & RULER 128K & Attention latency (ms) \\
\midrule
\multirow{9}{*}{1024} & \multirow{3}{*}{0.08} & 2 & 0.267 & 86.58 & 16147.8 \\
 &  & 4 & 0.365 & 87.28 & 18771.1 \\
 &  & 8 & 0.468 & 87.22 & 21951.6 \\
\cmidrule(lr){2-6}
 & \multirow{3}{*}{0.12} & 2 & 0.200 & 87.03 & 14790.7 \\
 &  & 4 & 0.280 & 87.26 & 16514.6 \\
 &  & 8 & 0.372 & 87.60 & 19732.4 \\
\cmidrule(lr){2-6}
 & \multirow{3}{*}{0.16} & 2 & 0.160 & 87.02 & 14355.9 \\
 &  & 4 & 0.225 & 87.86 & 15597.4 \\
 &  & 8 & 0.306 & 87.76 & 18390.4 \\
\midrule
\multirow{3}{*}{512} & \multirow{3}{*}{0.12} & 2 & 0.160 & 85.86 & 32084.0 \\
 &  & 4 & 0.229 & 86.37 & 31052.3 \\
 &  & 8 & 0.313 & 85.91 & 32260.9 \\
\bottomrule
\end{tabular}
}
\end{table}

\subsection{Zero-Copy Paged Execution}
\label{subsec:zerocopy}

CompactAttention uses FlashInfer~\cite{flashinfer} 0.6.9 as the paged attention execution backend. 
To support group-dependent block tables without copying K/V payloads, the accumulated KV cache is stored in a KV-head-major layout \([B, H_{kv}, L, D]\). 
Each \((\text{batch}, \text{KV head}, \text{block})\) triple then corresponds to a contiguous \([\text{block size}, D]\) memory region, which can be directly reinterpreted as a page without data movement.

To construct ragged page lists, the block-union mask is converted into CSR-style metadata \((\texttt{kv\_indptr}, \texttt{kv\_indices})\) by a fused CUDA kernel.
Only this metadata is passed to the paged attention backend; K/V payloads are never copied. 
In practice, we flatten the batch and KV-head dimensions into a pseudo-batch dimension and call the backend with \(\texttt{num\_kv\_heads}=1\), so that each pseudo-sequence carries its own page list.
For sub-KV-group union, query heads within the same execution group share the corresponding page list, enabling independent block tables for different execution groups without copying K/V payloads.

The current chunk is always kept fully open in the block mask.
If the current chunk were sparsified, causal masking would be applied in compacted-position space rather than original absolute-position space, breaking causal attention semantics.

\newpage

\section{Experiment Details}
\label{appendix:exp_details}

\subsection{Additional Latency Results}
\label{appendix:additional_latency}

\paragraph{LLaMA-3.1-8B-Instruct.}
Tables~\ref{tab:latency_rtx} and~\ref{tab:latency_h200} report raw attention and end-to-end latency measurements for LLaMA-3.1-8B-Instruct under chunked prefill.
Table~\ref{tab:latency_rtx} uses RTX PRO 6000 GPUs with TP=2, batch size 4, and chunk size 512, while Table~\ref{tab:latency_h200} uses H200 SXM GPUs with TP=2, batch size 8, and chunk size 1024.

Both CompactAttention variants reduce attention latency at long context lengths compared with their corresponding block-sparse baselines and dense attention, and these improvements also translate into end-to-end latency reductions.
At shorter context lengths, sparse methods can be slower in attention latency because pattern search and metadata construction overheads are not yet amortized, but their impact on end-to-end latency is smaller because attention accounts for a smaller fraction of total prefill time.

\begin{table}[h]
\centering
\caption{LLaMA-3.1-8B-Instruct attention and end-to-end latency (ms) across context lengths under chunked prefill with chunk size 512, measured on RTX PRO 6000 GPUs with TP=2 and batch size 4.}
\label{tab:latency_rtx}
\scalebox{0.9}{
\setlength{\tabcolsep}{10pt}
\renewcommand{\arraystretch}{0.95}
\begin{tabular}{lrrrrr}
\toprule
\multirow{2}{*}{Method} & \multicolumn{5}{c}{Context Length} \\
\cmidrule(lr){2-6}
& 8K & 16K & 32K & 64K & 128K \\
\midrule
\multicolumn{6}{c}{\textbf{Attention Latency}} \\
\midrule
Dense               & 423.1  & 1224.0 & 5213.3 & 17838.8 & 65971.1 \\
QUOKA               & 674.0  & 1599.5 & 5196.5 & 16325.4 & 56475.3 \\
XAttention          & 1095.5 & 2700.5 & 7770.5 & 24819.2 & 82617.0 \\
SeerAttention       & 1012.3 & 2548.7 & 6171.8 & 17078.2 & 49839.8 \\
\rowcolor{blue!8}
\textbf{CompactAttention}-SA & 1303.0 & 2675.7 & 5584.4 & 13823.5 & 32846.0 \\
FlashPrefill        & 496.6  & 1356.2 & 4547.7 & 14381.8 & 43750.5 \\
\rowcolor{blue!8}
\textbf{CompactAttention}-FP & 596.0  & 1393.6 & 3474.8 & 8945.9  & 24055.4 \\
\midrule
\multicolumn{6}{c}{\textbf{End-to-End Latency}} \\
\midrule
Dense               & 2299.7 & 4858.0 & 12421.8 & 32177.4 & 94944.3 \\
QUOKA               & 2470.4 & 5156.3 & 13116.8 & 30588.0 & 85375.3 \\
XAttention          & 2874.8 & 6243.9 & 14896.8 & 39206.6 & 113016.5 \\
SeerAttention       & 2911.3 & 6286.4 & 13620.0 & 32545.8 & 81774.4 \\
\rowcolor{blue!8}
\textbf{CompactAttention}-SA & 3139.8 & 6382.9 & 12706.7 & 28960.4 & 64126.7 \\
FlashPrefill        & 2294.2 & 4978.7 & 12088.4 & 29411.7 & 80108.5 \\
\rowcolor{blue!8}
\textbf{CompactAttention}-FP & 2345.1 & 4953.2 & 10691.7 & 23707.8 & 54130.6 \\
\bottomrule
\end{tabular}
}
\end{table}

\begin{table}[h]
\centering
\caption{LLaMA-3.1-8B-Instruct attention and end-to-end latency (ms) across context lengths under chunked prefill with chunk size 1024, measured on H200 SXM GPUs with TP=2 and batch size 8.}
\label{tab:latency_h200}
\scalebox{0.9}{
\setlength{\tabcolsep}{10pt}
\renewcommand{\arraystretch}{0.95}
\begin{tabular}{lrrrrr}
\toprule
\multirow{2}{*}{Method} & \multicolumn{5}{c}{Context Length} \\
\cmidrule(lr){2-6}
& 8K & 16K & 32K & 64K & 128K \\
\midrule
\multicolumn{6}{c}{\textbf{Attention Latency}} \\
\midrule
Dense               & 438.6  & 1294.7 & 4225.4 & 14882.5 & 55619.7 \\
QUOKA               & 582.3  & 1516.6 & 4629.7 & 15012.2 & 51292.4 \\
XAttention          & 923.9  & 2600.1 & 7729.7 & 24777.8 & 86821.4 \\
SeerAttention       & 1002.5 & 3073.4 & 9478.1 & 27632.5 & 85852.0 \\
\rowcolor{blue!8}
\textbf{CompactAttention}-SA & 780.2  & 1775.7 & 4393.8 & 10895.6 & 29468.1 \\
FlashPrefill        & 474.1  & 1279.6 & 3494.6 & 9451.1  & 24876.1 \\
\rowcolor{blue!8}
\textbf{CompactAttention}-FP & 533.6  & 1280.3 & 3160.1 & 7839.3  & 19899.9 \\
\midrule
\multicolumn{6}{c}{\textbf{End-to-End Latency}} \\
\midrule
Dense               & 1394.3 & 3160.7 & 7948.6  & 22327.7 & 70754.4 \\
QUOKA               & 1580.0 & 3571.4 & 8482.5  & 22451.8 & 66029.1 \\
XAttention          & 1877.0 & 4536.7 & 11649.8 & 32093.5 & 103870.2 \\
SeerAttention       & 1991.7 & 5029.2 & 13671.0 & 36406.8 & 105207.6 \\
\rowcolor{blue!8}
\textbf{CompactAttention}-SA & 1788.3 & 3745.1 & 8125.8  & 18453.3 & 45366.8 \\
FlashPrefill        & 1397.4 & 3135.1 & 7302.6  & 17392.8 & 41161.7 \\
\rowcolor{blue!8}
\textbf{CompactAttention}-FP & 1452.7 & 3128.3 & 6917.7  & 15634.6 & 35806.2 \\
\bottomrule
\end{tabular}
}
\end{table}

\newpage

\paragraph{Chunk Size Sensitivity.}
Table~\ref{tab:chunk_size_sensitivity} reports attention latency under different chunk sizes on LLaMA-3.1-8B-Instruct at 128K context length, measured on H200 SXM GPUs with TP=2 and batch size 8.
As chunk size increases, the number of chunked-prefill iterations decreases, reducing total attention latency.
CompactAttention-FP consistently improves latency across all chunk sizes.
Its relative speedup decreases as the chunk size grows because larger chunks increase Q-block union within each chunk, reducing effective sparsity, but it remains substantially faster than dense attention.

\begin{table}[h]
\centering
\caption{
Chunk-size sensitivity of attention latency on LLaMA-3.1-8B-Instruct at 128K context length, measured on H200 SXM GPUs with TP=2 and batch size 8.
}
\label{tab:chunk_size_sensitivity}
\scalebox{0.95}{
\renewcommand{\arraystretch}{0.95}
\setlength{\tabcolsep}{12pt}
\begin{tabular}{lccc}
\toprule
\multirow{2}{*}{Method} & \multicolumn{3}{c}{Chunk Size} \\
\cmidrule(lr){2-4}
& 512 & 1024 & 2048 \\
\midrule
\rowcolor{gray!8}
Dense & 72598.6 & 55619.7 & 47478.9 \\
\rowcolor{blue!8}
\textbf{CompactAttention}-FP & 25076.0 & 19899.9 & 19173.8 \\
\midrule
Speedup & 2.90$\times$ & 2.79$\times$ & 2.48$\times$ \\
\bottomrule
\end{tabular}
}
\end{table}

\paragraph{Attention Speedup on Qwen3-30B-A3B.}
Table~\ref{tab:qwen_speedup} reports additional attention speedup results on Qwen3-30B-A3B-Instruct-2507, a larger MoE model.
CompactAttention-FP uses sub-KV-group union with subgroup size 4, as described in Appendix~\ref{subsec:subkvgroup}.
The results show that CompactAttention-FP provides long-context latency gains as the accumulated KV cache grows: it is faster than FlashPrefill at all evaluated lengths and faster than dense attention from 64K onward.

\begin{table}[h]
  \centering
  \caption{Attention latency speedup on Qwen3-30B-A3B-Instruct-2507 under chunked prefill, measured on H200 SXM GPUs with TP=2, batch size 4, and chunk size 1024. MoE experts use a fused grouped-GEMM implementation.}
  \label{tab:qwen_speedup}
  \scalebox{0.95}{
  \setlength{\tabcolsep}{10pt}
  \begin{tabular}{lccc}
    \toprule
    Method & 32K & 64K & 128K \\
    \midrule
    Dense               & 1.00 & 1.00 & 1.00 \\
    FlashPrefill        & 0.82 & 0.99 & 1.24 \\
    \rowcolor{blue!8}
    \textbf{CompactAttention}-FP & 0.96 & 1.31 & 2.05 \\
    \bottomrule
  \end{tabular}
  }
\end{table}

\newpage

\textbf{Crossover across Chunk and Batch Sizes.} Tables~\ref{tab:h200_chunk_context} and~\ref{tab:h200_batch_context} report attention latency across chunk sizes and batch sizes. With batch size 8, CompactAttention-FP becomes faster than dense attention at 32K, 16K, and 16K for chunk sizes 512, 1024, and 2048, and faster than FlashPrefill at 32K, 32K, and 16K, respectively. With chunk size 1024, it does not overtake dense attention up to 128K at batch size 1, and overtakes it at 64K, 16K, and 16K for batch sizes 4, 8, and 16; against FlashPrefill, the crossover occurs at 64K, 64K, 32K, and 16K for batch sizes 1, 4, 8, and 16.

\begin{table}[h]
\centering
\caption{LLaMA-3.1-8B-Instruct attention latency (ms) across chunk sizes and context lengths under chunked prefill, measured on H200 SXM GPUs with TP=2 and batch size 8 (mean $\pm$ std over five runs).}
\label{tab:h200_chunk_context}
\scalebox{0.9}{
\setlength{\tabcolsep}{6pt}
\renewcommand{\arraystretch}{0.95}
\begin{tabular}{lrrrrr}
\toprule
\multirow{2}{*}{Method} & \multicolumn{5}{c}{Context Length} \\
\cmidrule(lr){2-6}
& 8K & 16K & 32K & 64K & 128K \\
\midrule
\multicolumn{6}{c}{\textbf{Chunk Size 512}} \\
\midrule
Dense                        & 564.3{\scriptsize$\pm$10.4} & 1628.3{\scriptsize$\pm$13.2} & 5349.4{\scriptsize$\pm$46.9} & 19156.4{\scriptsize$\pm$55.9} & 72598.6{\scriptsize$\pm$230.6} \\
FlashPrefill                 & 843.4{\scriptsize$\pm$32.9} & 1796.1{\scriptsize$\pm$20.7} & 4621.6{\scriptsize$\pm$13.9} & 11379.1{\scriptsize$\pm$70.6} & 30534.7{\scriptsize$\pm$283.9} \\
\rowcolor{blue!8}
\textbf{CompactAttention}-FP & 1152.2{\scriptsize$\pm$15.6} & 2181.8{\scriptsize$\pm$37.5} & 4350.9{\scriptsize$\pm$14.6} & 10473.0{\scriptsize$\pm$43.6} & 25076.0{\scriptsize$\pm$66.8} \\
\midrule
\multicolumn{6}{c}{\textbf{Chunk Size 1024}} \\
\midrule
Dense                        & 438.6{\scriptsize$\pm$1.6} & 1294.7{\scriptsize$\pm$0.7} & 4225.4{\scriptsize$\pm$11.0} & 14882.5{\scriptsize$\pm$17.9} & 55619.7{\scriptsize$\pm$34.0} \\
FlashPrefill                 & 474.1{\scriptsize$\pm$1.2} & 1279.6{\scriptsize$\pm$2.5} & 3494.6{\scriptsize$\pm$4.4} & 9451.1{\scriptsize$\pm$30.0} & 24876.1{\scriptsize$\pm$71.6} \\
\rowcolor{blue!8}
\textbf{CompactAttention}-FP & 533.6{\scriptsize$\pm$3.2} & 1280.3{\scriptsize$\pm$1.7} & 3160.1{\scriptsize$\pm$45.9} & 7839.3{\scriptsize$\pm$55.3} & 19899.9{\scriptsize$\pm$48.0} \\
\midrule
\multicolumn{6}{c}{\textbf{Chunk Size 2048}} \\
\midrule
Dense                        & 397.6{\scriptsize$\pm$1.6} & 1145.7{\scriptsize$\pm$5.3} & 3660.9{\scriptsize$\pm$15.2} & 12740.2{\scriptsize$\pm$19.4} & 47478.9{\scriptsize$\pm$30.9} \\
FlashPrefill                 & 438.4{\scriptsize$\pm$2.3} & 1194.5{\scriptsize$\pm$3.1} & 3250.2{\scriptsize$\pm$2.9} & 8650.9{\scriptsize$\pm$30.1} & 22106.5{\scriptsize$\pm$97.3} \\
\rowcolor{blue!8}
\textbf{CompactAttention}-FP & 446.6{\scriptsize$\pm$0.5} & 1139.3{\scriptsize$\pm$3.5} & 2891.1{\scriptsize$\pm$5.3} & 7376.0{\scriptsize$\pm$41.0} & 19173.8{\scriptsize$\pm$268.3} \\
\bottomrule
\end{tabular}
}
\end{table}

\begin{table}[h]
\centering
\caption{LLaMA-3.1-8B-Instruct attention latency (ms) across batch sizes and context lengths under chunked prefill, measured on H200 SXM GPUs with TP=2 and chunk size 1024 (mean $\pm$ std over five runs).}
\label{tab:h200_batch_context}
\scalebox{0.9}{
\setlength{\tabcolsep}{6pt}
\renewcommand{\arraystretch}{0.95}
\begin{tabular}{lrrrrr}
\toprule
\multirow{2}{*}{Method} & \multicolumn{5}{c}{Context Length} \\
\cmidrule(lr){2-6}
& 8K & 16K & 32K & 64K & 128K \\
\midrule
\multicolumn{6}{c}{\textbf{Batch Size 1}} \\
\midrule
Dense                        & 342.2{\scriptsize$\pm$3.9} & 626.3{\scriptsize$\pm$72.9} & 1392.9{\scriptsize$\pm$25.1} & 3160.5{\scriptsize$\pm$11.5} & 7715.4{\scriptsize$\pm$251.9} \\
FlashPrefill                 & 541.4{\scriptsize$\pm$2.4} & 1063.7{\scriptsize$\pm$55.7} & 2125.6{\scriptsize$\pm$238.6} & 5022.2{\scriptsize$\pm$166.2} & 10007.7{\scriptsize$\pm$565.4} \\
\rowcolor{blue!8}
\textbf{CompactAttention}-FP & 572.6{\scriptsize$\pm$26.3} & 1196.9{\scriptsize$\pm$49.1} & 2376.6{\scriptsize$\pm$40.8} & 4782.9{\scriptsize$\pm$43.6} & 9899.2{\scriptsize$\pm$68.7} \\
\midrule
\multicolumn{6}{c}{\textbf{Batch Size 4}} \\
\midrule
Dense                        & 250.1{\scriptsize$\pm$4.7} & 668.5{\scriptsize$\pm$6.7} & 2114.8{\scriptsize$\pm$4.5} & 7483.9{\scriptsize$\pm$19.4} & 27895.9{\scriptsize$\pm$19.4} \\
FlashPrefill                 & 435.1{\scriptsize$\pm$40.0} & 923.0{\scriptsize$\pm$59.5} & 2149.1{\scriptsize$\pm$3.2} & 5462.4{\scriptsize$\pm$13.0} & 14012.8{\scriptsize$\pm$120.9} \\
\rowcolor{blue!8}
\textbf{CompactAttention}-FP & 533.0{\scriptsize$\pm$32.9} & 1064.5{\scriptsize$\pm$26.0} & 2190.4{\scriptsize$\pm$87.9} & 4735.6{\scriptsize$\pm$28.4} & 11737.8{\scriptsize$\pm$73.2} \\
\midrule
\multicolumn{6}{c}{\textbf{Batch Size 8}} \\
\midrule
Dense                        & 438.6{\scriptsize$\pm$1.6} & 1294.7{\scriptsize$\pm$0.7} & 4225.4{\scriptsize$\pm$11.0} & 14882.5{\scriptsize$\pm$17.9} & 55619.7{\scriptsize$\pm$34.0} \\
FlashPrefill                 & 474.1{\scriptsize$\pm$1.2} & 1279.6{\scriptsize$\pm$2.5} & 3494.6{\scriptsize$\pm$4.4} & 9451.1{\scriptsize$\pm$30.0} & 24876.1{\scriptsize$\pm$71.6} \\
\rowcolor{blue!8}
\textbf{CompactAttention}-FP & 533.6{\scriptsize$\pm$3.2} & 1280.3{\scriptsize$\pm$1.7} & 3160.1{\scriptsize$\pm$45.9} & 7839.3{\scriptsize$\pm$55.3} & 19899.9{\scriptsize$\pm$48.0} \\
\midrule
\multicolumn{6}{c}{\textbf{Batch Size 16}} \\
\midrule
Dense                        & 875.2{\scriptsize$\pm$1.8} & 2606.9{\scriptsize$\pm$4.9} & 8501.6{\scriptsize$\pm$35.2} & 30680.5{\scriptsize$\pm$11.5} & OOM \\
FlashPrefill                 & 904.9{\scriptsize$\pm$1.6} & 2461.1{\scriptsize$\pm$8.0} & 6732.6{\scriptsize$\pm$9.3} & 17994.9{\scriptsize$\pm$11.5} & OOM \\
\rowcolor{blue!8}
\textbf{CompactAttention}-FP & 949.8{\scriptsize$\pm$3.1} & 2332.5{\scriptsize$\pm$7.1} & 5857.7{\scriptsize$\pm$34.5} & 14728.1{\scriptsize$\pm$37.3} & OOM \\
\bottomrule
\end{tabular}
}
\end{table}

\newpage
\clearpage
\subsection{Additional Accuracy Results}

\paragraph{RULER with Chunk Size 512.}
Table~\ref{tab:ruler_chunk512} reports RULER accuracy with chunk size 512 on LLaMA-3.1-8B-Instruct.
CompactAttention variants maintain accuracy close to dense attention across context lengths, with modest degradation at 128K.

\begin{table}[h]
\centering
\caption{\textbf{RULER accuracy across context lengths with chunk size 512.} 
We report Dense and CompactAttention variants on LLaMA-3.1-8B-Instruct.}
\label{tab:ruler_chunk512}
\scalebox{0.95}{
\renewcommand{\arraystretch}{1.0}
\setlength{\tabcolsep}{8pt}
\begin{tabular}{lcccc}
\toprule
\multirow{2}{*}{Method} & \multicolumn{4}{c}{LLaMA-3.1-8B-Instruct} \\
\cmidrule(lr){2-5}
& 32K & 64K & 128K & Avg \\
\midrule
\rowcolor{gray!8}
Dense & 88.23 & 83.52 & 76.59 & 82.78 \\
\rowcolor{blue!8}
\textbf{CompactAttention}-SA & 89.24 & 83.47 & 74.56 & 82.42 \\
\rowcolor{blue!8}
\textbf{CompactAttention}-FP & 88.26 & 83.59 & 73.48 & 81.78 \\
\bottomrule
\end{tabular}
}
\end{table}

\paragraph{LongBench V2.}
Table~\ref{tab:longbenchv2_accuracy} reports the full LongBench V2 breakdown for all methods on LLaMA-3.1-8B-Instruct. 
As discussed in Section~\ref{subsec:accuracy}, QUOKA shows consistent degradation across categories compared to dense attention, while block-sparse methods largely preserve accuracy. 
Both CompactAttention-SA and CompactAttention-FP maintain accuracy comparable to dense attention and their corresponding block-sparse counterparts across all difficulty levels and context lengths.

\begin{table}[h]
\centering
\caption{LLaMA-3.1-8B-Instruct LongBench V2 accuracy comparison across task categories with chunk size 1024.}
\label{tab:longbenchv2_accuracy}
\scalebox{0.95}{
\renewcommand{\arraystretch}{0.95}
\setlength{\tabcolsep}{8pt}
\begin{tabular}{lrrrrrr}
\toprule
Method & Overall & Easy & Hard & Short & Medium & Long \\
\midrule
\rowcolor{gray!8}
Dense         & 30.4 & 32.3 & 29.3 & 37.2 & 24.7 & 30.6 \\
QUOKA         & 26.8 & 29.2 & 25.4 & 33.3 & 21.9 & 25.9 \\
XAttention    & 28.2   & 30.2   & 27.0   & 35.6   & 24.2   & 24.1   \\
SeerAttention & 29.4 & 29.7 & 29.3 & 34.4 & 24.2 & 31.5 \\
\rowcolor{blue!8}
\textbf{CompactAttention}-SA         & 29.6 & 31.2 & 28.6 & 36.7 & 23.7 & 29.6 \\
FlashPrefill  & 28.0   & 28.6   & 27.7   & 33.9   & 22.8   & 28.7   \\
\rowcolor{blue!8}
\textbf{CompactAttention}-FP         & 30.4 & 31.2 & 29.9 & 37.2 & 24.2 & 31.5 \\
\bottomrule
\end{tabular}
}
\end{table}


\subsection{Copy-Based vs. Zero-Copy Execution}
\label{appendix:metadata}

Table~\ref{tab:copy_metadata_batch_scaling} extends the execution-strategy ablation in Figure~\ref{fig:ablation}(c) across batch sizes.
CompactAttention-FP (Copy) materializes selected KV blocks into a compact buffer and then invokes dense attention directly with FlashAttention-2.
Its metadata overhead includes the preprocessing required to determine where selected KV blocks should be copied in the compact buffer.
In contrast, CompactAttention-FP eliminates explicit KV copying and represents selected KV blocks as paged-attention metadata.
Its metadata overhead includes per-execution-group block-table construction and the \texttt{plan()} overhead of the FlashInfer wrapper.
The results show that copy overhead in CompactAttention-FP (Copy) grows with batch size, whereas the metadata overhead of CompactAttention-FP grows much more slowly, supporting the benefit of zero-copy paged execution in batched chunked prefill.

\begin{table}[h]
\centering
\caption{
Batch-size scaling of copy and metadata overhead for CompactAttention-FP.
Latency is measured in milliseconds.
}
\label{tab:copy_metadata_batch_scaling}
\scalebox{0.92}{
\setlength{\tabcolsep}{8pt}
\begin{tabular}{lccc|cc}
\toprule
\multirow{2}{*}{Batch Size}
& \multicolumn{3}{c|}{\textbf{CompactAttention}-FP (Copy)} 
& \multicolumn{2}{c}{\textbf{CompactAttention}-FP} \\
\cmidrule(lr){2-4} \cmidrule(lr){5-6}
& Metadata & Copy & Computation & Metadata & Computation \\
\midrule
1  & 0.20 & 0.30 & 1.65  & 0.29 & 1.54  \\
2  & 0.13 & 0.52 & 2.79  & 0.24 & 2.96  \\
4  & 0.13 & 1.01 & 5.13  & 0.31 & 4.48  \\
8  & 0.13 & 1.99 & 9.59  & 0.47 & 8.33  \\
16 & 0.16 & 3.97 & 21.26 & 0.89 & 15.73 \\
\bottomrule
\end{tabular}
}
\end{table}

Table~\ref{tab:gather_zerocopy} compares the two execution strategies at the attention
level on H200: with the same selected KV blocks, CompactAttention-FP (Copy) already
achieves 2.62$\times$ speedup over dense attention at 128K, and zero-copy execution
raises this to 2.79$\times$, showing that most of the gain comes from reducing the
executed KV set while zero-copy removes the remaining copy cost.

\begin{table}[h]
\centering
\caption{Attention latency (ms) of copy-based and zero-copy execution with the same selected KV blocks on LLaMA-3.1-8B-Instruct, measured on H200 SXM GPUs with TP=2, batch size 8, and chunk size 1024.}
\label{tab:gather_zerocopy}
\scalebox{0.9}{
\setlength{\tabcolsep}{10pt}
\begin{tabular}{lrrr}
\toprule
\multirow{2}{*}{Method} & \multicolumn{3}{c}{Context Length} \\
\cmidrule(lr){2-4}
& 32K & 64K & 128K \\
\midrule
Dense                               & 4225.4 & 14882.5 & 55619.7 \\
FlashPrefill                        & 3494.6 & 9451.1  & 24876.1 \\
\textbf{CompactAttention}-FP (Copy) & 3320.6 & 8223.6  & 21189.6 \\
\rowcolor{blue!8}
\textbf{CompactAttention}-FP        & 3160.1 & 7839.3  & 19899.9 \\
\bottomrule
\end{tabular}
}
\end{table}

\subsection{Comparison with FlexPrefill}
\label{app:flexprefill}

We evaluate FlexPrefill~\cite{flexprefill} as an additional baseline under chunked
prefill. Since the official implementation supports only batch size 1 and equal query
and key lengths, we extend it to the rectangular attention of chunked prefill without
changing its pattern search, and use the hyperparameters from the original paper.
As shown in Table~\ref{tab:flexprefill}, FlexPrefill maintains competitive accuracy
but is 2.99--4.81$\times$ slower than dense attention, as its fine-grained pattern
search is repeated at every chunk over the growing KV cache.

\begin{table}[h]
  \centering
  \caption{Comparison with FlexPrefill on LLaMA-3.1-8B-Instruct (chunk size 1024). Latency is measured on H200 with TP=2 and batch size 8.}
  \label{tab:flexprefill}
  \small
  \begin{tabular}{lcccccc}
    \toprule
    \multirow{2}{*}{Method} & \multicolumn{3}{c}{RULER Accuracy} & \multicolumn{3}{c}{Attention Latency (ms)} \\
    \cmidrule(lr){2-4} \cmidrule(lr){5-7}
    & 32K & 64K & 128K & 32K & 64K & 128K \\
    \midrule
    Dense               & 88.23 & 83.52 & 76.59 & 4225.4  & 14882.5 & 55619.7  \\
    FlexPrefill         & 89.42 & 82.96 & 73.30 & 20339.1 & 56731.9 & 166356.5 \\
    \rowcolor{blue!8}
    \textbf{CompactAttention}-FP & 88.77 & 83.96 & 74.17 & 3160.1  & 7839.3  & 19899.9  \\
    \bottomrule
  \end{tabular}
\end{table}

\newpage

\subsection{Post-Union Sparsity Analysis}
\label{app:sparsity}

Keep ratios in this section are compute-weighted: kept and total KV blocks are summed
over chunk iterations before taking the ratio, counting mandatory recent blocks,
including the current chunk, in both. Medians, P90 values, and the fraction of
near-dense executions (keep ratio $\geq 0.9$) are weighted by KV length over individual
(chunk, layer, execution group) executions.

Table~\ref{tab:keep_stages} decomposes the keep ratio of CompactAttention-FP by
aggregation step. Q-block union and intra-group union each increase the keep ratio,
and larger chunks increase it further. Near-dense executions become rare as the
context grows, and CompactAttention-FP remains faster than both dense attention and
FlashPrefill in all settings. Table~\ref{tab:keep_tasks} shows that the post-union
keep ratio is similar across RULER tasks. Table~\ref{tab:qwen_layer} shows that on
Qwen3-30B-A3B the keep ratio varies widely across layers and execution groups, and
that sub-KV-group union keeps it lower than full KV-group union.

\begin{table}[h]
  \centering
  \caption{Compute-weighted keep ratio after each aggregation step and attention speedup of CompactAttention-FP on LLaMA-3.1-8B-Instruct (RULER VT). Speedup is measured on H200 with TP=2 and batch size 8.}
  \label{tab:keep_stages}
  \small
  \setlength{\tabcolsep}{4pt}
  \begin{tabular}{llcccccccc}
    \toprule
    \multirow{2}{*}{Context} & \multirow{2}{*}{Chunk} & \multirow{2}{*}{Pre-union} & \multirow{2}{*}{Q-block union} & \multicolumn{4}{c}{Group union} & \multicolumn{2}{c}{Speedup over} \\
    \cmidrule(lr){5-8} \cmidrule(lr){9-10}
    & & & & Mean & Median & P90 & $\geq 0.9$ & Dense & FlashPrefill \\
    \midrule
    \multirow{3}{*}{32K}  & 512  & 0.281 & 0.406 & 0.584 & 0.538 & 0.994 & 0.206 & 1.23$\times$ & 1.06$\times$ \\
                          & 1024 & 0.274 & 0.477 & 0.652 & 0.634 & 1.000 & 0.277 & 1.34$\times$ & 1.11$\times$ \\
                          & 2048 & 0.260 & 0.552 & 0.717 & 0.732 & 1.000 & 0.352 & 1.27$\times$ & 1.12$\times$ \\
    \midrule
    \multirow{3}{*}{64K}  & 512  & 0.179 & 0.260 & 0.415 & 0.333 & 0.887 & 0.095 & 1.83$\times$ & 1.09$\times$ \\
                          & 1024 & 0.174 & 0.308 & 0.468 & 0.397 & 0.949 & 0.125 & 1.90$\times$ & 1.21$\times$ \\
                          & 2048 & 0.168 & 0.364 & 0.525 & 0.467 & 0.993 & 0.159 & 1.73$\times$ & 1.17$\times$ \\
    \midrule
    \multirow{3}{*}{128K} & 512  & 0.100 & 0.150 & 0.260 & 0.184 & 0.589 & 0.035 & 2.90$\times$ & 1.22$\times$ \\
                          & 1024 & 0.098 & 0.182 & 0.303 & 0.227 & 0.671 & 0.049 & 2.79$\times$ & 1.25$\times$ \\
                          & 2048 & 0.096 & 0.223 & 0.354 & 0.281 & 0.754 & 0.064 & 2.48$\times$ & 1.15$\times$ \\
    \bottomrule
  \end{tabular}
\end{table}

\begin{table}[h]
  \centering
  \caption{Compute-weighted post-union keep ratio of CompactAttention-FP across RULER tasks on LLaMA-3.1-8B-Instruct (chunk size 1024).}
  \label{tab:keep_tasks}
  \small
  \begin{tabular}{lcccc}
    \toprule
    Context & VT & NIAH-1 & CWE & QA-1 \\
    \midrule
    32K  & 0.652 & 0.675 & 0.651 & 0.699 \\
    64K  & 0.468 & 0.469 & 0.422 & 0.494 \\
    128K & 0.303 & 0.299 & 0.269 & 0.351 \\
    \bottomrule
  \end{tabular}
\end{table}

\begin{table}[h]
  \centering
  \caption{Compute-weighted post-union keep ratio per execution group on Qwen3-30B-A3B-Instruct-2507 (128K, chunk size 1024) for the three layers with the highest and lowest average keep ratio under subgroup size 4. The last column reports the layer average under full KV-group union (subgroup size 8).}
  \label{tab:qwen_layer}
  \small
  \setlength{\tabcolsep}{4pt}
  \begin{tabular}{lccccccccc|c}
    \toprule
    \multirow{2}{*}{Layer} & \multicolumn{9}{c|}{Subgroup size 4} & Subgroup size 8 \\
    \cmidrule(lr){2-10}
    & G0 & G1 & G2 & G3 & G4 & G5 & G6 & G7 & Avg. & Avg. \\
    \midrule
    2  & 0.654 & 0.438 & 0.303 & 0.367 & 0.873 & 0.428 & 0.505 & 0.508 & 0.510 & 0.636 \\
    0  & 0.758 & 0.973 & 0.324 & 0.393 & 0.626 & 0.311 & 0.206 & 0.081 & 0.459 & 0.598 \\
    38 & 0.319 & 0.534 & 0.574 & 0.466 & 0.465 & 0.161 & 0.291 & 0.466 & 0.410 & 0.490 \\
    \midrule
    7  & 0.218 & 0.166 & 0.123 & 0.133 & 0.165 & 0.123 & 0.097 & 0.096 & 0.140 & 0.159 \\
    5  & 0.105 & 0.177 & 0.168 & 0.115 & 0.120 & 0.145 & 0.080 & 0.159 & 0.134 & 0.161 \\
    8  & 0.051 & 0.065 & 0.029 & 0.170 & 0.129 & 0.114 & 0.096 & 0.095 & 0.094 & 0.118 \\
    \bottomrule
  \end{tabular}
\end{table}



\end{document}